\def\year{2020}\relax
\documentclass[letterpaper]{article}
\usepackage{aaai20}
\usepackage{times}
\usepackage{helvet}
\usepackage{courier}
\usepackage{url}
\usepackage{graphicx}
\usepackage{amsmath}
\usepackage{amsthm}
\usepackage{amssymb}
\usepackage{amsfonts}
\usepackage{algorithm}
\usepackage{algorithmic}
\usepackage{booktabs}
\usepackage{multirow}
\usepackage{eqparbox}
\usepackage{array}

\title{Adversarial Domain Adaptation with Domain Mixup}
\author{
	Minghao Xu,\textsuperscript{\rm 1}
	Jian Zhang,\textsuperscript{\rm 1}
	Bingbing Ni,\textsuperscript{\rm 1,2}\footnotemark[1]
	Teng Li,\textsuperscript{\rm 3}\\
	\bf \Large Chengjie Wang,\textsuperscript{\rm 4}
	Qi Tian,\textsuperscript{\rm 5}
	Wenjun Zhang\textsuperscript{\rm 1}\\
	\textsuperscript{\rm 1}Shanghai Jiao Tong University, China,
	\textsuperscript{\rm 2}Huawei Hisilicon,\\
	\textsuperscript{\rm 3}Anhui University,
	\textsuperscript{\rm 4}Youtu Lab, Tencent,
	\textsuperscript{\rm 5}Huawei Noah’s Ark Lab\\
	\{xuminghao118, stevenash0822, nibingbing, zhangwenjun\}@sjtu.edu.cn, nibingbing@hisilicon.com,\\
	tenglwy@gmail.com, jasoncjwang@tencent.com, tian.qi1@huawei.com
}

\begin{document}

\maketitle

\begin{abstract}

Recent works on domain adaptation reveal the effectiveness of adversarial learning on filling the discrepancy between source and target domains. However, two common limitations exist in current adversarial-learning-based methods. First, samples from two domains alone are not sufficient to ensure domain-invariance at most part of latent space. Second, the domain discriminator involved in these methods can only judge real or fake with the guidance of hard label, while it is more reasonable to use soft scores to evaluate the generated images or features, \emph{i.e.}, to fully utilize the inter-domain information. In this paper, we present adversarial domain adaptation with \textbf{domain mixup} (DM-ADA), which guarantees domain-invariance in a more continuous latent space and guides the domain discriminator in judging samples' difference relative to source and target domains. Domain mixup is jointly conducted on pixel and feature level to improve the robustness of models. Extensive experiments prove that the proposed approach can achieve superior performance on tasks with various degrees of domain shift and data complexity.

\end{abstract}

\section{Introduction} \label{section1}
In recent years\footnotetext[0]{*The corresponding author is Bingbing Ni.}, with the appearance of Convolutional Neural Networks (CNNs), many classification-based challenges have been tackled with an extremely high accuracy. These powerful CNN architectures, like AlexNet \cite{alexnet} and ResNet \cite{resnet}, are capable of efficiently extracting low-level and high-level features with the guidance of labeled data. However, because of the existence of domain shift, models trained on a specific domain suffer from poorer performance when transferred to another domain. This problem is of vital significance in the case where labeled data are unavailable on target domain. Thus how to use these unlabeled data from target domain to fill the domain discrepancy is the main issue in the field of domain adaptation \cite{survey}. 

Beginning with the work of gradient reverse \cite{gradient_reverse}, a group of domain adaptation methods based on adversarial learning were proposed. In the research of this subfield, a domain discriminator is introduced to judge the domain attribute on feature or image level. In order to fool this domain discriminator, the extracted features should be domain-invariant, which is the basic motivation of adversarial domain adaptation. However, just like most variants of Generative Adversarial Networks (GANs) \cite{gan}, the domain discriminator is \emph{only} guided by the hard label information and \emph{rarely} explores the intrinsic structure of data distribution. Namely, each data point that shifts from both domains should be judged by a latent soft label, \emph{i.e.}, a probability value, instead of a hard assignment of ``1" or ``0". In addition, the distribution of domain-invariant latent vectors is fitted using limited patterns of source and target features, \emph{i.e.}, the interaction of features from two domains has not been considered to enrich feature patterns. 

\begin{figure}[t]
    \centering
    \includegraphics[width=.96\columnwidth]{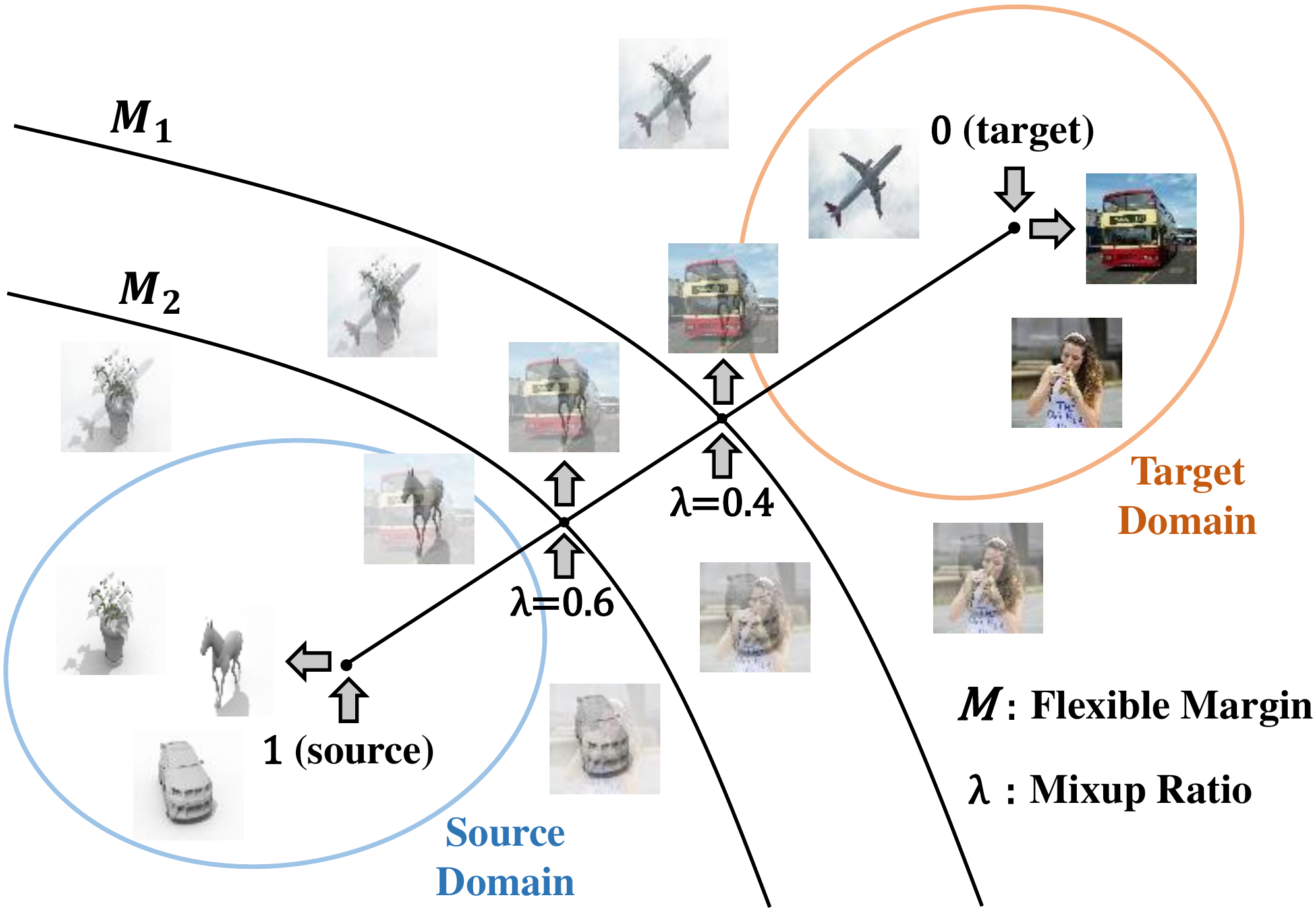}
    \caption{The mixup samples are obtained by linearly interpolating between source and target domain images with mixup ratio $\lambda$. The domain discriminator learns how to output soft scores for generated images with the guidance of mixup images and \emph{soft labels} (mixup ratios). Also, a flexible margin is learned through comparing mixup samples' difference relative to two domains. (Best viewed in color.)}\label{fig_motivation}
\end{figure}

In this paper, inspired by VAE-GAN \cite{vae-gan}, we develop a generative-adversarial-based framework to simultaneously train the classification network and generate auxiliary source-like images from learned embeddings of both domains. On the basis of this framework, \textbf{domain mixup} on pixel and feature level is proposed to alleviate two existing drawbacks. (1) Just as shown in Figure \ref{fig_motivation}, we would like to instruct the domain discriminator to explore the intrinsic structure of source and target distributions through triplet loss with a flexible margin and applying domain classification with mixed images and corresponding soft labels (\emph{i.e.}, mixup ratio), which provides abundant intermediate status between two separate domains. (2) In order to expand the searching range in latent space, linear interpolations of source and target features are exploited. Since the subsequent nonlinear neural network can easily ruin the linear mixed information, the extracted features of mixed images are not used for augmentation directly. This operation leads to a more continuous domain-invariant latent distribution, which benefits the performance on target domain when the oscillation of data distribution occurs in the test phase.

We evaluate the image recognition performance of our approach on three benchmarks with different extent of domain shift. Experiments prove the effectiveness of our approach, and we achieve state-of-the-art in most settings. The contributions of our work are summarized as follows:

\begin{itemize}
	\item We design an adversarial training framework which maps both domains to a common latent distribution, and efficiently transfer our knowledge learned on the supervised domain to its unsupervised counterpart. 
	\item Domain mixup on pixel and feature level accompanied with well-designed soft domain labels is proposed to improve the generalization ability of models. This method promotes the generalization ability of feature extractor and obtains a domain discriminator judging samples' difference relative to two domains with refined scores.
	\item We extensively evaluate our approach under different settings, and our approach achieves superior results even when the domain shift is high and the data distribution is complex. 
\end{itemize}

\section{Related Work} \label{section2}
Domain adaptation is a frequently used technique to promote the generalization ability of models trained on a single domain in many Computer Vision tasks. In this section, we describe existing domain adaptation methods and compare our approach with them.

The transferability of Deep Neural Networks is proved in \cite{deep_transfer_ability}, and deep learning methods for domain adaptation can be classified into several categories. Maximum Mean Discrepancy (MMD) \cite{kernel_two_sample,ddc} is a way to measure the similarity of two distributions. Weighted Domain Adaptation Network (WDAN) \cite{wdan} defines the weighted MMD with class conditional distribution on both domains. The multiple kernel version of MMD (MK-MMD) is explored in \cite{Long2015} to define the distance between two distributions. In addition, specific deep neural networks are constructed to restrict the domain-invariance of top layers by aligning the second-order statistics \cite{deep_coral}.

Adversarial Training \cite{domain_adversarial,domain_adversarial_journal} is another way to transfer domain information. RevGrad \cite{gradient_reverse} designs a double branched architecture for object classification and domain classification respectively. Adversarial Discriminative Domain Adaptation (ADDA) \cite{adda} trains two feature extractors for source and target domains respectively, and produces embeddings fooling the discriminator. Other works optimize the performance on target domain by capturing complex multimode structures \cite{multi-adversarial,cdan}, exploring task-specific decision boundaries \cite{max_discrepancy,adversarial_dropout,joint_pixel_feature}, aligning the attention regions \cite{adversarial_attention} and applying structure-aware alignment \cite{gcan}. In addition, the label information of target domain is explored in recent works \cite{collaborative,semantic,gcan}.

Another group of methods perform adaptation by applying adversarial loss on pixel level. Source domain images are adapted as if they are drawn from target domain using generative adversarial networks in \cite{pixel_level,co-gan}, and generated samples expand the training set. Furthermore, image generation and training the task-specific classifier are accomplished simultaneously in \cite{deep_reconstruction,generate_to_adapt}. Cycle-consistency is also considered in \cite{cycada} to enforce the consistency of relevant semantics.

\textbf{Comparison with existing GAN-based approaches.} Although former works use GAN as a manner of data augmentation \cite{pixel_level,co-gan} or producing domain adaptive gradient information \cite{deep_reconstruction,generate_to_adapt}, they may be trapped in the mismatch between generated data and assigned hard labels. We further explore the usage of \textbf{domain mixup} on pixel and feature level to enhance the robustness of adaptation models. On one hand, pixel-level mixup prompts the domain discriminator to excavate the intrinsic structure of source and target distributions. On the other hand, feature-level mixup facilitates a more continuous feature distribution in the latent space with low domain shift. 

\begin{figure*}[t]
    \centering
    \includegraphics[width=.97\textwidth]{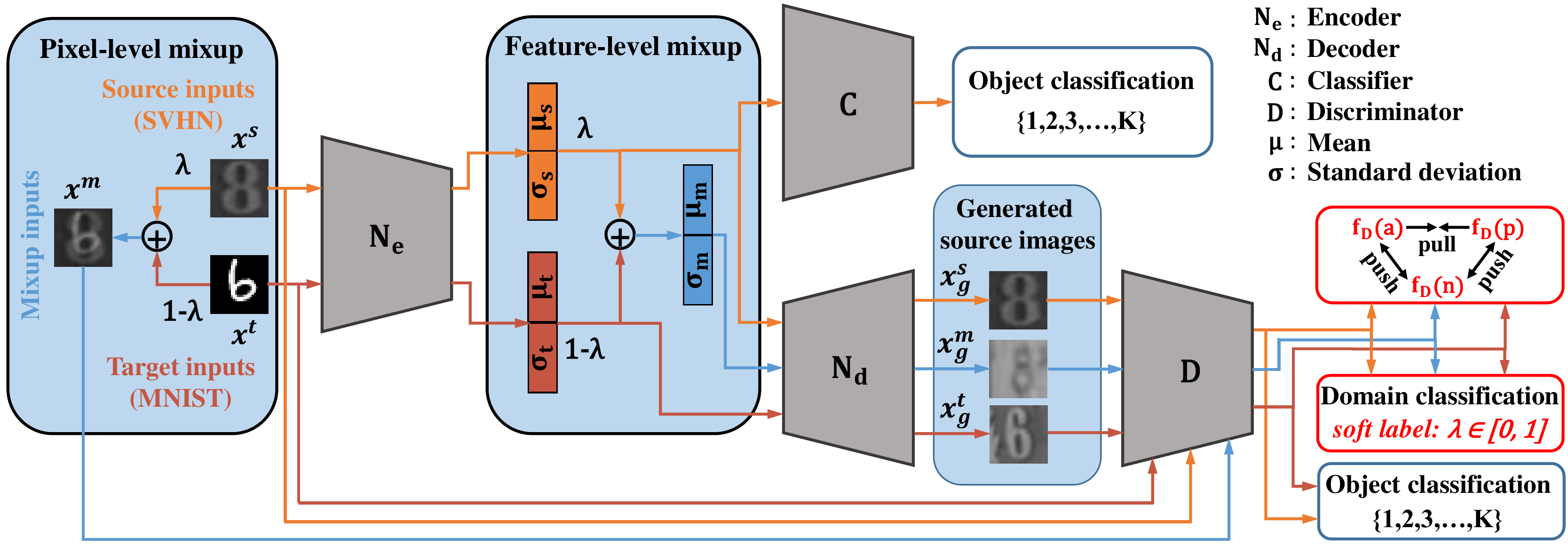}
	\caption{Illustration of the pipeline in the training phase. First, mixup inputs $x^m$ are obtained by mixing two domains' inputs $x^s$ and $x^t$. Encoder $N_e$ maps source and target inputs to $(\mu_s, \sigma_s)$ and $(\mu_t, \sigma_t)$ respectively. In the latent space, feature embeddings of two domains are mixed to produce mixup features $(\mu_m, \sigma_m)$. After that, the framework is split to two branches. On one branch, classifier $C$ performs $K$-way object classification. On the other branch, latent codes are decoded by $N_d$, and the min-max game between $N_d$ and $D$ facilitates domain-invariance on category level. (Best viewed in color.)}\label{fig_model}
\end{figure*}

\section{Adversarial Domain Adaptation \\ with Domain Mixup} \label{section3}
In unsupervised domain adaptation, a source domain dataset $(X_s, Y_s) $ $ = \{(x_i^s,y_i^s)\}_{i=1}^{n_s}$ with $n_s$ labeled samples and a target domain dataset $X_t = \{x_j^t\}_{j=1}^{n_t}$ with $n_t$ unlabeled samples are available. It is assumed that source samples $x^s$ obey the source distribution $P_s$, and target samples $x^t$ obey the target distribution $P_t$. In addition, both domains share the same label space $Y=\{1, 2, 3, \cdots, K\}$, where $K$ is the number of classes. 

\subsection{Framework of DM-ADA} \label{section3_1}
In this work, a variant of VAE-GAN \cite{vae-gan} is applied to the domain adaptation task. Figure \ref{fig_model} presents an overview of the whole framework. For the input, there are three kinds: source domain images, target domain images and mixup images obtained by pixel-wise addition of source and target images. Just as conventional variational autoencoder \cite{vae}, an encoder $N_e$ maps inputs from source and target domains to the standard Gaussian distribution $\mathcal{N}(0,\textbf{I})$. For every sample, a mean vector $\mu$ and a standard deviation vector $\sigma$ are served as the feature embedding. On feature level, the feature embeddings of two domains are also linearly mixed to produce mixup features $(\mu_m, \sigma_m)$. After that, the framework is split into two branches. For one branch, the embedding of source domain is used to do $K$-way object classification by the classifier $C$. For the other branch, source and target domain are aligned on category level through enforcing the decoded images to be source-like and preserve class information of inputs. Details are stated in the following parts.

\textbf{Domain mixup on two levels.} To explore the internal structure of data from two domains, source domain images $x^s$ and target domain images $x^t$ are linearly interpolated \cite{mixup} to produce mixup images $x^m$ and corresponding soft domain labels $l^{m}_{dom}$ as follows:
\begin{equation} \label{eq_mixup_1}
    x^m = \lambda x^s + (1-\lambda) x^t ,
\end{equation}
\begin{equation} \label{eq_mixup_2}
    l^{m}_{dom} = \lambda l^{s}_{dom} + (1-\lambda) l^{t}_{dom} = \lambda ,
\end{equation}
where $\lambda \in [0, 1]$ is the mixup ratio, and $\lambda \sim \textrm{Beta}(\alpha, \alpha)$, in which $\alpha$ is constantly set as 2.0 in all experiments. $l^{s}_{dom}$ and $l^{t}_{dom}$ represent the domain label of source and target data, which are manually set as 1 and 0. 

Inputs of source and target domains are then embedded to $(\mu_s, \sigma_s)$ and $(\mu_t, \sigma_t)$ in the latent space by a shared encoder $N_e$.  
In order to yield a more continuous domain-invariant latent distribution, two domains' embeddings are linearly mixed to produce mixup feature embedding $(\mu_m, \sigma_m)$:
\begin{equation} \label{eq_feature_mixup_1}
    \mu_m = \lambda \mu_s + (1-\lambda) \mu_t ,
\end{equation}
\begin{equation} \label{eq_feature_mixup_2}
    \sigma_m = \lambda \sigma_s + (1-\lambda) \sigma_t ,
\end{equation}
where $\lambda$ equals to the one used in pixel-level mixup.

\textbf{Restricting encoder with priori.} Just as conventional VAE \cite{vae}, the encoder $N_e$ is regularized by a standard gaussian priori over the latent distribution. The objective is to narrow the Kullback-Leibler divergence between posteriori and priori:
\begin{equation}
    \min \limits_{N_e} \mathcal{L}_{KL} ,
\end{equation}
\begin{equation}
    \mathcal{L}_{KL} = D_{KL} \big ( \mathcal{N}(\mu, \sigma) || \mathcal{N}(0,\textbf{I}) \big ) ,
\end{equation}
where $\mu$ and $\sigma$ are the encoded mean and standard deviation of source and target images.

\textbf{Supervised training for classifier.} The classifier $C$ is optimized with cross entropy loss defined on source domain, and the objective is as follows:
\begin{equation}
    \min \limits_{N_e, C} \mathcal{L}_{C} ,
\end{equation}
\begin{equation}
    \mathcal{L}_{C} = -\mathbf{E}_{x^s \sim P_s} \sum_{i=1}^{K} y^s_i \, \textrm{log} \big ( C([\mu_s, \sigma_s]) \big ) ,
\end{equation}
where $[\cdot]$ denotes concatenation. It is worth noticing that classifier $C$ can't be replaced by the object classification branch of discriminator $D$, since the adapted features are only passed directly to $C$, which enhances $C$'s performance on target domain.

\textbf{Decoding latent codes.} Before the generation phase, we first define the one-hot object class label $l_{cls}$ and a one-dimensional uncertainty compensation $l_{comp}$ for both domains and mixup features as below:
\begin{equation}
\begin{aligned}
l_{cls}^s = [0,0,\cdots,1,\cdots,0], & \quad l_{comp}^s = 0 , \\
l_{cls}^t = [0,0,\cdots,0,\cdots,0], & \quad l_{comp}^t = 1 ,  \\
l_{cls}^m = [0,0,\cdots,\lambda,\cdots,0], & \quad l_{comp}^m = 1 - \lambda ,
\end{aligned}
\end{equation}
where $1$ and $\lambda$ are on the $y^s$-th position of $l_{cls}^s$ and $l_{cls}^m$ to indicate the known class label for both features respectively. For all features derived from target domain or mixup procedure, since the class labels remain uncertain, $l_{comp}$ is set as a compensation to normalize the sum of vector $l_{cls}$ and $l_{comp}$ to 1. 
After that, decoder $N_d$ predicts the auxiliary generated images $x_g$ as below:
\begin{equation}
    x_g = N_d([\mu, \sigma, z, l_{cls}, l_{comp}]) ,
\end{equation}
where $z$ is the noise vector randomly sampled from standard Gaussian distribution.

\textbf{Adversarial domain alignment.} Compared with previous adversarial-learning-based methods \cite{gradient_reverse,adda,multi-adversarial}, we constrain domain-invariance not only on source and target domains, but also on the intermediate representations between two domains. The min-max optimization objective on different domains are defined as follows:

\begin{equation}
    \min \limits_{N_e, N_d} \max \limits_{D} \ \mathcal{L}^s_{adv} + \mathcal{L}^t_{adv} + \mathcal{L}^m_{adv} ,
\end{equation}

\begin{equation}
    \mathcal{L}^s_{adv} = \mathbf{E}_{x^s \sim P_s} \, \textrm{log} \big ( D_{dom} (x^s) \big ) + \textrm{log} \big ( 1 - D_{dom} (x^s_g) \big ) ,
\end{equation}

\begin{equation}
    \mathcal{L}^t_{adv} = \mathbf{E}_{x^t \sim P_t} \, \textrm{log} \big ( 1 - D_{dom} (x^t_g) \big ) ,
\end{equation}

\begin{equation}
    \mathcal{L}^{m}_{adv} = \mathbf{E}_{x^s \sim P_s, x^t \sim P_t} \, \textrm{log} \big ( 1 - D_{dom} (x^m_g) \big ) ,
\end{equation}
where $D_{dom}$ is the domain classification branch of $D$. During training process, the mixup features can well be mapped to somewhere in-between source and target domain on pixel level, and it is more proper to assign them with scores between 0 and 1. Domain classification loss $\mathcal{L}^m_{soft}$ is utilized to guide domain discriminator output such soft scores:
\begin{equation}
    \min \limits_{D} \mathcal{L}^m_{soft} ,
\end{equation}
\begin{equation}
\begin{aligned}
\mathcal{L}^m_{soft} = & -\mathbf{E}_{x^s \sim P_s, x^t \sim P_t}  \, l_{dom}^m \textrm{log} \big ( D_{dom} (x^m) \big ) \\
& + (1 - l_{dom}^m) \textrm{log} \big ( 1 - D_{dom} (x^m) \big ) .
\end{aligned}
\end{equation}

We further introduce a triplet loss $\mathcal{L}^m_{tri}$ to constrain mixup samples' distance to source and target domains, which makes domain discriminator easier to converge:
\begin{equation}
    \min \limits_{D} \mathcal{L}^m_{tri} ,
\end{equation}
\begin{equation}
\begin{aligned}
    \mathcal{L}^m_{tri} = & \mathbf{E}_{x^s \sim P_s, x^t \sim P_t} \, \big [ ||f_D(a) - f_D(p)||_2^2 \\ & - ||f_D(a) - f_D(n)||_2^2 + f_{tri}(\lambda) \big ]_{+} ,
\end{aligned}
\end{equation}
where $f_D$ is the feature extractor of $D$, and $[\cdot]_{+} = max(0, \cdot)$ denotes the hinge loss function; $(a,p,n) = (x^m, x^s, x^t)$, when $\lambda \geqslant 0.5$, and $(a,p,n) = (x^m, x^t, x^s)$, otherwise. Considering that samples with more source or target domain components should have larger difference with the counterpart domain, a flexible margin $f_{tri}(\lambda) = |2\lambda - 1|$ is used.

\textbf{Category-level domain alignment.} In order to ensure the identical categories' features of two domains are mapped nearby in the latent space, classification loss $\mathcal{L}^s_{cls}$ and $\mathcal{L}^t_{cls}$ are introduced to ensure the class-consistency between decoded images and inputs:

\begin{equation}
    \min \limits_{N_e, N_d, D} \mathcal{L}^s_{cls} + \mathcal{L}^t_{cls} ,
\end{equation}
\begin{equation}
    \mathcal{L}^s_{cls} = -\mathbf{E}_{x^s \sim P_s} \sum_{i=1}^{K} y^s_i \, \textrm{log} \big ( D_{cls}(x^s_g) \big ) ,
\end{equation}
\begin{equation}
    \mathcal{L}^t_{cls} = -\mathbf{E}_{x^t \sim P_t} \sum_{i=1}^{K} \hat{y}^t_i \, \textrm{log} \big ( D_{cls}(x^t_g) \big ) ,
\end{equation}
where $D_{cls}$ is the object classification branch of $D$, and $\hat{y}^t$ is the pseudo label estimated by classifier $C$. So as to eliminate falsely labeled samples which harm domain adaptation, we filter out those samples whose classification confidence below a certain threshold $\tau$. Considering the fact that domain discrepancy is gradually filled along training, $\tau$ is adaptively adjusted following the strategy in \cite{collaborative}. 

\begin{algorithm}[t]  
	\caption{ Training procedure of DM-ADA }  
	\label{training_algorithm}  
	\begin{algorithmic}
		\STATE {\bfseries Input:}  
		Source domain: $X_s$ and $Y_s$, target domain: $X_t$ and the number of iterations $N$.
		\STATE {\bfseries Output:}
		Configurations of DM-ADA.
	    \STATE {\bfseries Initialize} $\omega$ and $\varphi$
	    \STATE {\bfseries Initialize} $\theta_{N_e}$, $\theta_{N_d}$, $\theta_{C}$ and $\theta_{D}$
	    \FOR{$n=1$ {\bfseries to} $N$}
	    \STATE $(x^s, y^s) \gets$ RANDOMSAMPLE$(X_s, Y_s)$
	    \STATE $(x^t) \gets$ RANDOMSAMPLE$(X_t)$
	    \STATE $\lambda \gets$ RANDOMSAMPLE$(\textrm{Beta}(\alpha, \alpha))$
	    \STATE $(x^m, l_{dom}^m) \gets$ Eq. (\ref{eq_mixup_1}, \ref{eq_mixup_2}) \COMMENT{\small Get mixup images}
	    \STATE $(\mu, \sigma) \gets N_e(x)$  \COMMENT{\small Get feature embeddings}
	    \STATE $(\mu_m, \sigma_m) \gets$ Eq. (\ref{eq_feature_mixup_1}, \ref{eq_feature_mixup_2}) \COMMENT{\small Get mixup features}
	    \STATE $x_g = N_d(\mu, \sigma, z, l_{cls}, l_{comp})$ \COMMENT{\small Generate images}
	    \STATE {\bfseries Optimize} four subnetworks $D$, $N_d$, $C$ and $N_e$ in turn:
	    \begin{small}
	    \begin{displaymath}
	    \begin{aligned}
	    \theta_{D} \stackrel{+}{\gets} & - \nabla_{\theta_{D}} \big ( \mathcal{L}^s_{cls} + \omega (\mathcal{L}^m_{tri} + \mathcal{L}^m_{soft}) \\
	    & + \varphi (\mathcal{L}^s_{adv} + \mathcal{L}^t_{adv} + \mathcal{L}^m_{adv}) \big ) \\
	    \theta_{N_d} \stackrel{+}{\gets} & -\nabla_{\theta_{N_d}} (\mathcal{L}^s_{cls} - \varphi \mathcal{L}^s_{adv} ) \\
	    \theta_{C} \stackrel{+}{\gets} & -\nabla_{\theta_{C}} (\mathcal{L}_{C}) \\
	    \theta_{N_e} \stackrel{+}{\gets} & -\nabla_{\theta_{N_e}} \big ( \mathcal{L}_{C} + \mathcal{L}^s_{cls} + \mathcal{L}^t_{cls} \\
	    & + \omega \mathcal{L}_{KL} - \varphi (\mathcal{L}^t_{adv} + \mathcal{L}^m_{adv}) \big )
	    \end{aligned}
	    \end{displaymath}
	    \end{small}
	    \ENDFOR
	\end{algorithmic}  
\end{algorithm}

\subsection{Training Procedure} \label{section3_2}

The proposed iterative training procedure is summarized in Algorithm \ref{training_algorithm}. In each iteration, the input source and target samples are first mixed on pixel level to instruct the domain discriminator to output soft labels. After the samples of two domains are mapped to the latent space, their embeddings are mixed to produce mixup features. The images generated on the basis of these feature embeddings are constrained to be source-like and preserve inputs' class information, so that the latent distribution is facilitated to be domain-invariant and discriminative. In all experiments, we set $\alpha$ as 2.0, since domain mixup can't effectively explore the linear space between two domains when the value of $\alpha$ is small, and more analysis of $\alpha$ can be found in supplementary material. $\omega$ and $\varphi$ are hyper-parameters that trade off among losses with different orders of magnitude. According to the after sensitivity analysis, the adaptation performance of our approach is not too sensitive to the value of $\omega$ and $\varphi$, and these hyper-parameters share the same value among different tasks.

\subsection{Discussion} \label{section3_3}
\textbf{Pixel-level domain mixup.} The work of \cite{mixup} proposes the \emph{mixup} vicinal distribution as a manner to encourage the model to behave linearly in-between training examples. Another work \cite{autoencoder_interpolation} improves interpolation's continuity in latent space and benefits downstream tasks. In adversarial domain adaptation, we also would like to lead the domain discriminator to behave linearly between source and target domains. As a result, the domain discriminator is of high capacity to accurately judge the generated images containing oscillations to two domains. In our implementation, such discriminator is trained with pairs of linearly mixed image $x^m = \lambda x^s + (1-\lambda) x^t$ and corresponding soft label $l^m_{dom} = \lambda$, where $x^m$ simulates an oscillation mode to two domains and $\lambda$ provides the guidance. Combined with feature-level mixup, pixel-level mixup can further narrow the domain discrepancy, which is shown in the after ablation study. 

\begin{figure}[t]
    \centering
    \includegraphics[width=.98\columnwidth]{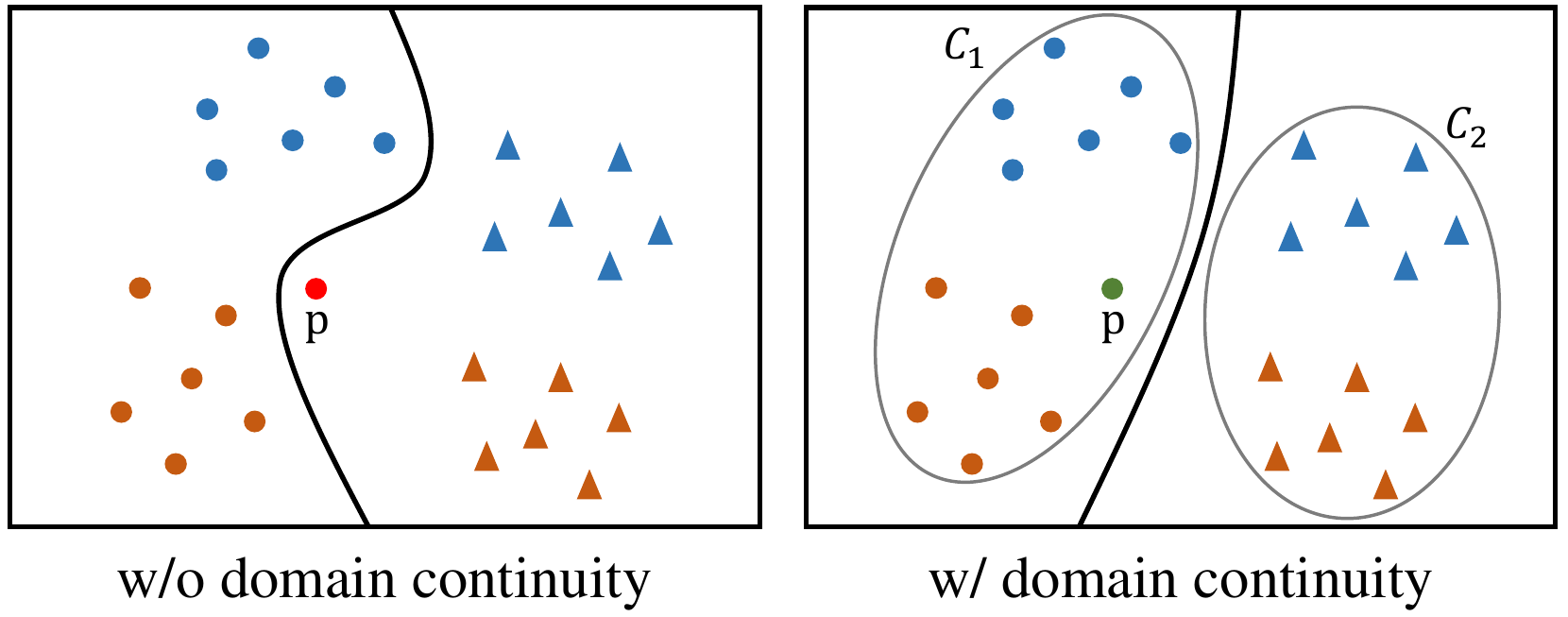}
    \caption{The comparison between with and without domain continuity. Circle \& triangle: two classes; Blue \& orange: source and target domain; $p$: test sample. (Best viewed in color.)}\label{fig_feature_mixup}
\end{figure}

\textbf{Feature-level domain mixup.} Existing works attempt to map source and target domains to a common latent distribution, while limited data can not guarantee most parts of the latent space domain-invariant. In order to yield a more continuous domain-invariant latent distribution, the mixup features of two domains are exploited. 

We use an intuitive example to illustrate the effectiveness of domain continuity on aligning source and target domains. As shown in Figure \ref{fig_feature_mixup}, the biased test sample $p$ may be misclassified without the constraint of domain continuity. However, through adding the mixup feature embedding to the training process, the latent codes between the same class of two domains should also be domain-invariant, which forms the intra-class clusters $C_1$ and $C_2$. Thus the decision boundary is refined, and the biased samples in these clusters can be classified correctly.

\section{Experiments} \label{section4}

In this section, we first introduce the experimental setup. Then, the classification performance on three domain adaptation benchmarks are presented. Finally, ablation study and sensitivity analysis are conducted for the proposed approach. 

\begin{table}[t]
\centering
\small
\caption{Classification accuracy (mean $\pm$ std \%) values of target domain over five independent runs on the digits datasets. The best performance is indicated in \textbf{bold} and the second best one is \underline{underlined}.}\label{table_digit}
\resizebox{.98\columnwidth}{!}{
\begin{tabular}{ccccc}
    \toprule[1.0pt]
    \multirow{2}{*}{Method} & MN $\rightarrow$ & MN $\rightarrow$ & US $\rightarrow$ & SV $\rightarrow$ \\
    & US (p) & US (f) & MN & MN \\
    \hline
    Source only & $76.0 \pm 1.8$ & $79.3 \pm 0.7$ & $59.5 \pm 1.9$ & $62.1 \pm 1.2$ \\
    \hline
    MMD \shortcite{Long2015} & \rule[1mm]{0.2cm}{0.15mm} & $81.1 \pm 0.3$ & \rule[1mm]{0.2cm}{0.15mm} & $71.1 \pm 0.5$ \\
    RevGrad \shortcite{gradient_reverse} & $77.1 \pm 1.8$ & $85.1 \pm 0.8$ & $73.0 \pm 2.0$ & $73.9 \pm 1.2$ \\
    CoGAN \shortcite{co-gan} & $91.2 \pm 0.8$ & \rule[1mm]{0.2cm}{0.15mm} & $89.1 \pm 1.0$ & \rule[1mm]{0.2cm}{0.15mm} \\
    DRCN \shortcite{deep_reconstruction} & $91.8 \pm 0.1$ & \rule[1mm]{0.2cm}{0.15mm} & $73.7 \pm 0.0$ & $82.0 \pm 0.1$ \\
    ADDA \shortcite{adda} & $89.4 \pm 0.2$ & \rule[1mm]{0.2cm}{0.15mm} & $90.1 \pm 0.8$ & $76.0 \pm 1.8$ \\
    PixelDA \shortcite{pixel_level} & \rule[1mm]{0.2cm}{0.15mm} & $95.9 \pm 0.7$ & \rule[1mm]{0.2cm}{0.15mm} & \rule[1mm]{0.2cm}{0.15mm} \\
    MSTN \shortcite{semantic} & $92.9 \pm 1.1$ & \rule[1mm]{0.2cm}{0.15mm} & \rule[1mm]{0.2cm}{0.15mm} & $91.7 \pm 1.5$ \\
    GTA \shortcite{generate_to_adapt} & $92.8 \pm 0.9$ & $95.3 \pm 0.7$ & $90.8 \pm 1.3$ & $92.4 \pm 0.9$ \\
    ADR \shortcite{adversarial_dropout} & $\underline{93.2} \pm 2.5$ & $\underline{96.1} \pm 0.3$ & $\underline{93.1} \pm 1.3$ & $\underline{95.0} \pm 1.9$ \\
    \hline
    DM-ADA (ours) & $\textbf{94.8} \pm 0.7$ & $\textbf{96.7} \pm 0.5$ & $\textbf{94.2} \pm 0.9$ & $\textbf{95.5} \pm 1.1$ \\
    \bottomrule[1.0pt]
\end{tabular}}
\end{table}

\begin{table*}[t]
\centering
\small
\caption{Classification accuracy (mean $\pm$ std \%) values of target domain over five independent runs on the Office-31 dataset. The best performance is indicated in \textbf{bold} and the second best one is \underline{underlined}.}\label{table_office}
\resizebox{.97\textwidth}{!}{
\begin{tabular}{cccccccc}
    \toprule[1.0pt]
    Method & A $\rightarrow$ W & D $\rightarrow$ W & W $\rightarrow$ D & A $\rightarrow$ D & D $\rightarrow$ A & W $\rightarrow$ A & Average \\
    \hline
    {AlexNet (source only) \shortcite{alexnet}} & $60.6 \pm 0.4$ & {$95.4 \pm 0.2$} & {$99.0 \pm 0.1$} & {$64.2 \pm 0.3$} & {$45.5 \pm 0.5$} & {$48.3 \pm 0.5$} & {68.8} \\
    \hline
    {TCA \shortcite{tca}} & {$59.0 \pm 0.0$} & {$90.2 \pm 0.0$} & {$88.2 \pm 0.0$} & {$57.8 \pm 0.0$} & {$51.6 \pm 0.0$} & {$47.9 \pm 0.0$} & {65.8} \\
    {DDC \shortcite{ddc}} & {$61.0 \pm 0.5$} & {$95.0 \pm 0.3$} &
    {$98.5 \pm 0.3$} & {$64.9 \pm 0.4$} & {$47.2 \pm 0.5$} & {$49.4 \pm 0.4$} & {69.3} \\
    {DAN \shortcite{Long2015}} & {$68.5 \pm 0.3$} & {$96.0 \pm 0.1$} & {$99.0 \pm 0.1$} & {$66.8 \pm 0.2$} & {$50.0 \pm 0.4$} & {$49.8 \pm 0.3$} & {71.7} \\
    {RevGrad \shortcite{gradient_reverse}} & {$73.0 \pm 0.5$} & {$96.4 \pm 0.3$} & {$99.2 \pm 0.3$} & {$72.3 \pm 0.3$} & {$52.4 \pm 0.4$} & {$50.4 \pm 0.5$} & {74.1} \\
    {DRCN \shortcite{deep_reconstruction}} & {$68.7 \pm 0.3$} & {$96.4 \pm 0.3$} & {$99.0 \pm 0.2$} & {$66.8 \pm 0.5$} & {$56.0 \pm 0.5$} & {$54.9 \pm 0.5$} & 73.6 \\
    {MADA \shortcite{multi-adversarial}} & {$78.5 \pm 0.2$} & {$\textbf{99.8} \pm 0.1$} & {$\textbf{100} \pm 0.0$} & {$74.1 \pm 0.1$} & {$56.0 \pm 0.2$} & {$54.5 \pm 0.3$} & {77.1} \\
    {MSTN \shortcite{semantic}} & {$80.5 \pm 0.4$} & {$96.9 \pm 0.1$} & {$\underline{99.9} \pm 0.1$} & {$74.5 \pm 0.4$} & {$62.5 \pm 0.4$} & {$60.0 \pm 0.6$} & {79.1} \\
    {GCAN \shortcite{gcan}} & {$\underline{82.7} \pm 0.1$} & {$\underline{97.1} \pm 0.1$} & {$99.8 \pm 0.1$} & {$\underline{76.4} \pm 0.5$} & {$\textbf{64.9} \pm 0.1$} & {$\underline{62.6} \pm 0.3$} & {\underline{80.6}} \\
    \hline
    {DM-ADA (ours)} & {$\textbf{83.9} \pm 0.4$} & {$\textbf{99.8} \pm 0.1$} & {$\underline{99.9} \pm 0.1$} & {$\textbf{77.5} \pm 0.2$} & {$\underline{64.6} \pm 0.4$} & {$\textbf{64.0} \pm 0.5$} & {\textbf{81.6}} \\
    \bottomrule[1.0pt]
\end{tabular}}
\end{table*}

\subsection{Experimental Setup} \label{section4_1}
In this part, we describe the network architectures and hyper-parameters of different tasks. Our approach is implemented with PyTorch deep learning framework \cite{pytorch}.

\textbf{Digits experiments.} In this part of experiments, we construct four subnetworks with train-from-scratch architectures following \cite{generate_to_adapt}. Four Adam optimizers with base learning rate 0.0004 are utilized to optimize these submodels for 100 epochs. The hyper-parameters $\omega$ and $\varphi$ are set as 0.1 and 0.01 respectively, and their values are constant in all experiments. All of the input images of encoder and discriminator are resized to $32 \times 32$.

\textbf{Office experiments.} For the encoder, the last layer of AlexNet \cite{alexnet} is replaced with two parallel fully connected layers producing 256 dimensional vectors respectively, and former layers are initialized with the model pretrained on ImageNet \cite{imagenet}. The encoder is fine-tuned with base learning rate 0.0001 for 100 epochs, and the base learning rate of other three submodels is set as 0.001. The inputs of encoder and discriminator are resized to $227 \times 227$ and $64 \times 64$ respectively.

\textbf{VisDA experiments.} ResNet-101 \cite{resnet} serves as the base architecture, and it is initialized with the model pretrained on ImageNet \cite{imagenet}. The learning rate setting is same as that in the office experiments, and the results are reported after 20 epochs training. The inputs of encoder and discriminator are resized to $224 \times 224$ and $64 \times 64$ respectively.

\subsection{Classification on Digits Datasets} \label{section4_2}

\textbf{Dataset.} In this set of experiments, three digits datasets are used: MNIST \cite{mnist}, USPS \cite{usps} and Street View House Numbers (SVHN) \cite{svhn}. Each dataset contains ten classes corresponding to number 0 to 9. Four settings are used for measurement: MN $\rightarrow$ US (p): sampling 2000 images from MNIST and 1800 images from USPS; MN $\rightarrow$ US (f) and US $\rightarrow$ MN: using the full training set of MNIST and USPS; SV $\rightarrow$ MN: using the full training set of SVHN and MNIST. 

\textbf{Results.} Table \ref{table_digit} presents the results of our approach in comparison with other adaptation approaches on the digits datasets. For the source only test, we use the same encoder and classifier architectures as the ones used in our approach. The reported results are averaged over five independent runs with random initialization. Our approach achieves the state-of-the-art performance on all four settings. Especially, it outperforms former GAN-based approaches \cite{pixel_level,deep_reconstruction,generate_to_adapt}, which illustrates the effectiveness of the proposed architecture on aligning source and target domains.

\begin{table}[t]
\centering
\small
\caption{Classification accuracy on the validation set of VisDA-2017 challenge.}\label{table_visda}
\resizebox{.90\columnwidth}{!}{
\begin{tabular}{cc}
    \toprule[1.0pt]
    Method & Accuracy (\%) \\
    \hline
    ResNet-101 (source only) \shortcite{resnet} & 52.4 \\
    \hline
    RevGrad \shortcite{gradient_reverse} & 57.4 \\
    DAN \shortcite{Long2015} & 62.8 \\
    JAN \shortcite{Long2017} & 65.7 \\
    GTA \shortcite{generate_to_adapt} & 69.5 \\
    MCD-DA \shortcite{max_discrepancy} & 71.9 \\
    ADR \shortcite{adversarial_dropout} & \underline{73.5} \\
    \hline
    DM-ADA (ours) & \textbf{75.6} \\
    \bottomrule[1.0pt]
\end{tabular}}
\end{table}

\subsection{Classification on Office-31} \label{section4_3}

\begin{figure*}[t]
    \centering
    \includegraphics[width=.98\textwidth]{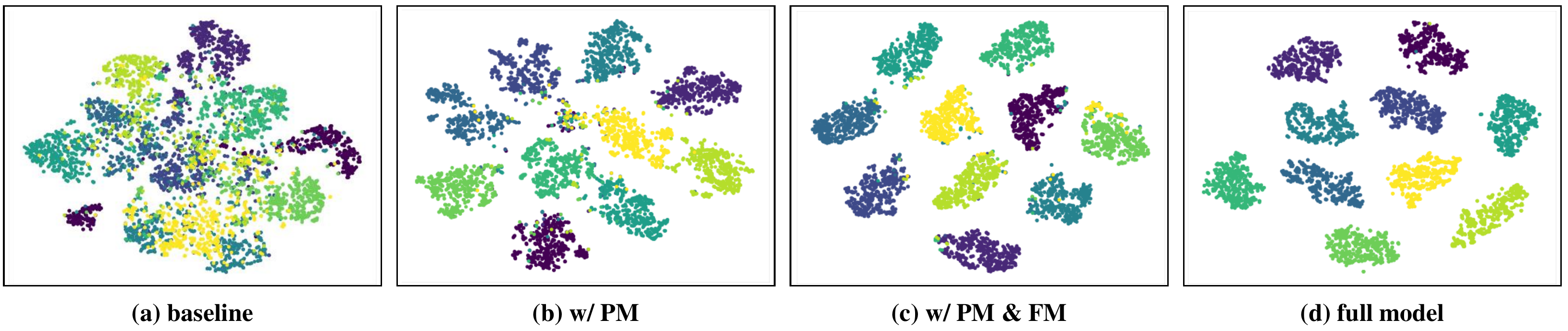}
	\caption{The t-SNE \cite{tsne} visualization of target domain's feature distribution on the transfer task SVHN $\rightarrow$ MNIST under four different configurations. (Best viewed in color.)}\label{fig_tsne}
\end{figure*}

\textbf{Dataset.} Office-31 \cite{office} is a standard domain adaptation benchmark commonly used in previous researches. Three distinct domains, Amazon(A), Webcam(W) and DSLR(D), make up of the whole Office-31 dataset. Each domain contains the same 31 classes of office supplies. All transfer tasks of three domains are used for evaluation. 

\textbf{Results.} Table \ref{table_office} reports the performance of our method compared with other works. The results of AlexNet trained with only source domain data serves as the lower bound. Our approach obtains the best performance in three of four hard cases: A $\rightarrow$ W, W $\rightarrow$ A and A $\rightarrow$ D. For two easier cases: W $\rightarrow$ D and D $\rightarrow$ W, our approach achieves accuracy higher than 99.5\% and ranks the first two places. Given the fact that the number of samples per class is limited in the Office-31 dataset, our approach manages to improve the performance by providing augmented samples and features.

\subsection{Classification on VisDA-2017} \label{section4_4}

\begin{table}[t]
\centering
\small
\caption{Effectiveness of pixel-level mixup (PM), feature-level mixup (FM) and triplet loss (Tri).} \label{table_ablation_1}
\resizebox{.88\columnwidth}{!}{
\begin{tabular}{ccc|cc}
    \toprule[1.0pt]
    PM & FM & Tri & $\mathcal{A}$-distance & Accuracy (\%) \\
    \hline
    & & & 1.528 & 76.7 \\
    \checkmark & & & 1.519 & 78.1 \\
    \checkmark & & \checkmark & 1.508 & 79.4 \\
    & \checkmark & & 1.497 & 82.1 \\
    \checkmark & \checkmark & & 1.492 & 83.2 \\
    \checkmark & \checkmark & \checkmark & 1.489 & 83.9 \\
    \bottomrule[1.0pt]
\end{tabular}}
\end{table}

\begin{table}[t]
\centering
\small
\caption{Effectiveness of $D_{cls}$ and pseudo target labels.} \label{table_ablation_2}
\resizebox{.88\columnwidth}{!}{
\begin{tabular}{cc|cc}
    \toprule[1.0pt]
    $D_{cls}$ & pseudo & $\mathcal{A}$-distance & Accuracy (\%) \\
    \hline
    & & 1.503 & 80.6 \\
    \checkmark & & 1.496 & 82.3 \\
    \checkmark & \checkmark & 1.489 & 83.9 \\
    \bottomrule[1.0pt]
\end{tabular}}
\end{table}

\textbf{Dataset.} The VisDA-2017 \cite{visda} challenge proposes a large-scale dataset for visual domain adaptation. The training domain is composed of synthetic renderings of 3D models. The validation domain is made up of photo-realistic images drawn from MSCOCO \cite{coco}. Both domains contain the same 12 classes of objects. 

\textbf{Results.} Table \ref{table_visda} reports the results on the VisDA-2017 cross-domain classification dataset. The ResNet-101 model pretrained on ImageNet acts as the baseline. Our approach achieves the highest accuracy among all adaptation approaches, and exceeds the baseline with a great margin. Under the condition that large domain shift exists, like transferring from synthetic objects to real images in this task, we think that the triplet loss and soft label play a critical role in excavating intermediate status between two domains.

\subsection{Ablation Study} \label{section4_5}

\textbf{Metrics.} Two metrics are employed. (1) $\mathcal{A}$-distance \cite{A_distance,A_distance_2} serves as a measure of cross-domain discrepancy. Inputted with extracted features of two domains, a SVM classifier is used to classify the source and target domain features, and the generalization error is defined as $\epsilon$. Then the $\mathcal{A}$-distance can be calculated as: $d_{\mathcal{A}} = 2(1 - 2 \epsilon)$.
(2) Classification accuracy on target domain serves as a measure of task-specific performance. In this part of experiments, both metrics are evaluated on the task A $\rightarrow$ W.

\begin{figure}[t]
    \centering
    \includegraphics[width=.97\columnwidth]{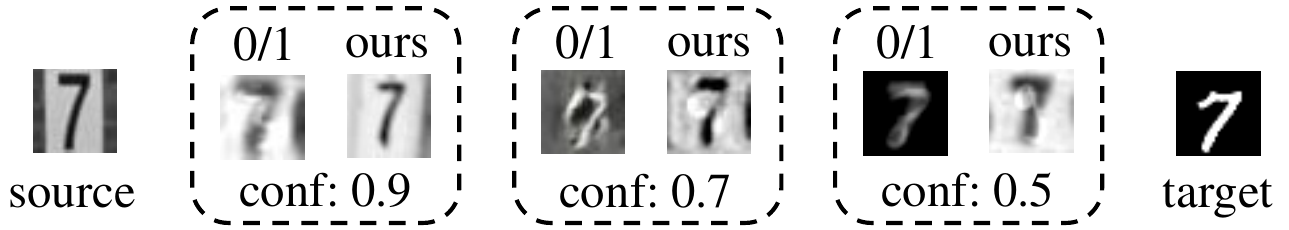}
    \caption{The generated images with confidence 0.9, 0.7 and 0.5 using 0/1 and our discriminator. (task: SV $\rightarrow$ MN)} \label{fig_discriminator}
\end{figure}

\textbf{Effect of pixel-level and feature-level mixup.} Table \ref{table_ablation_1} examines the effectiveness of pixel-level mixup (PM) and feature-level mixup (FM). The first row only uses the images and feature embeddings from two domains for training, and it serves as the baseline. In the fourth row, feature-level mixup achieves notable improvement compared with baseline, since the domain-invariant latent space is facilitated to be more continuous in this configuration. In the fifth row, pixel-level mixup further enhance model's performance through guiding discriminator output soft scores between 0 and 1, which means it is an essential auxiliary scheme for feature-level mixup. In Figure \ref{fig_discriminator}, compared with traditional 0/1 discriminator, our discriminator leads to more source-like generated images, which means the domain discrepancy can be further narrowed via pixel-level mixup. 

\textbf{Effect of triplet loss.} In Table \ref{table_ablation_1}, we evaluate another key component, \emph{i.e.}, triplet loss (Tri). In the third and sixth rows, it can also be observed that model's performance is improved after adding the triplet loss to discriminator's training process, since this loss ease the convergence of domain discriminator. We further utilize t-SNE \cite{tsne} to visualize the feature distribution of target domain on the task SV $\rightarrow$ MN. As shown in Figure \ref{fig_tsne}, the features of different classes are separated most clearly in the full model, \emph{i.e.}, with domain mixup on two levels and triplet loss.

\textbf{Effect of $D_{cls}$ and pseudo target labels.} In order to conduct category-aware alignment between source and target domains, the classification branch of discriminator $D_{cls}$ and pseudo target labels are employed, and the effectiveness of them is examined in Table \ref{table_ablation_2}. After appending $D_{cls}$, classification accuracy increases by 1.7\%, since this branch facilitates generated images to preserve the class information contained in inputs, which makes domain adaptation perform on the same categories of two domains. On such basis, pseudo target labels introduce the discriminative information of target domain to the adaptation process and make model's performance state-of-the-art.

\begin{figure}[t]
    \centering
    \includegraphics[width=.97\columnwidth]{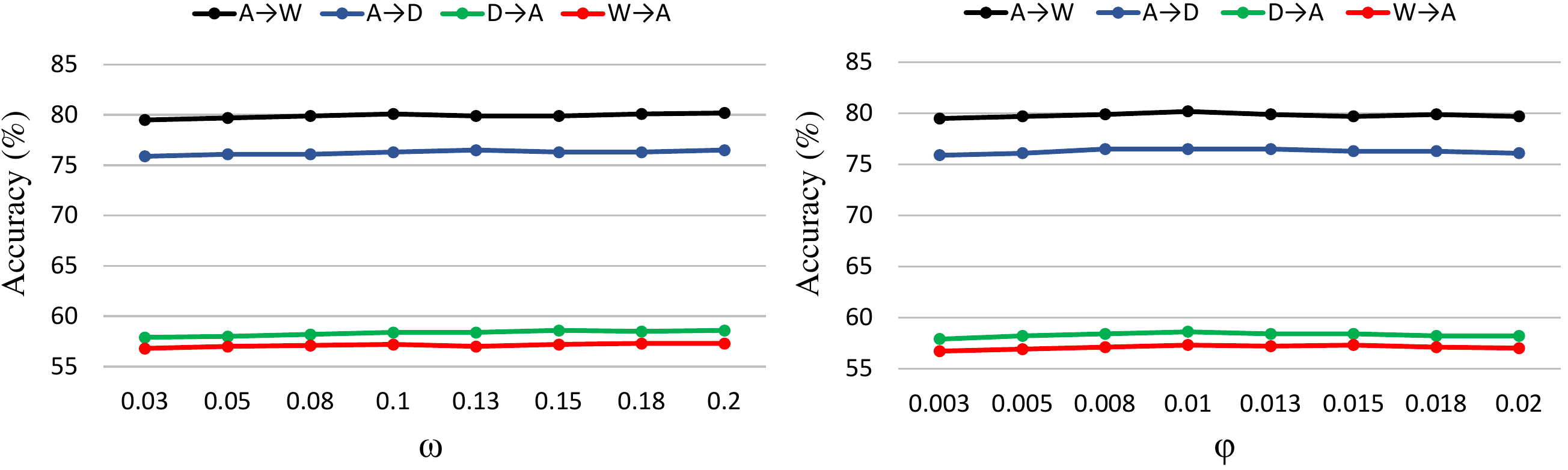}
    \caption{The sensitivity analysis of $\omega$ (left) and $\varphi$ (right). Four hard-to-transfer tasks of Office-31 dataset are involved.} \label{fig_sensitivity}
\end{figure}

\subsection{Sensitivity Analysis} \label{section4_6}

In this section, we discuss our approach's sensitivity to hyper-parameters $\omega$ and $\varphi$ which trade off among losses with different orders of magnitude. Four hard-to-transfer tasks of Office-31 dataset are used for evaluation. In Figure \ref{fig_sensitivity}, it can be observed that the transfer performance is not sensitive to the variance of $\omega$ and $\varphi$ near 0.1 and 0.01, respectively. In consequence, we can set $\omega$ and $\varphi$ as 0.1 and 0.01 for all tasks, and the transfer performance should be satisfactory.

\section{Conclusion} \label{section5}
In this paper, we address the problem of unsupervised domain adaptation. A GAN-based architecture is constructed to transfer knowledge from source domain to target domain. In order to facilitate a more continuous domain-invariant latent space and fully utilize the inter-domain information, we propose the domain mixup on pixel and feature level. Extensive experiments on adaptation tasks with different extent of domain shift and data complexity demonstrate the predominant performance of our approach.

\section{Acknowledge} \label{section6}
This work was supported by National Science Foundation of China (61976137, U1611461). This work was also supported by SJTU-BIGO Joint Research Fund, and CCF-Tencent Open Fund. 

\newpage
\small
\bibliographystyle{aaai}
\bibliography{reference.bib} 

\begin{thebibliography}{}

\bibitem[\protect\citeauthoryear{Ajakan \bgroup et al\mbox.\egroup
  }{2013}]{domain_adversarial}
Ajakan, H.; Germain, P.; Larochelle, H.; Laviolette, F.; and Marchand, M.
\newblock 2013.
\newblock Domain-adversarial neural networks.
\newblock {\em CoRR} abs/1412.4446.

\bibitem[\protect\citeauthoryear{Ben{-}David \bgroup et al\mbox.\egroup
  }{2010}]{A_distance}
Ben{-}David, S.; Blitzer, J.; Crammer, K.; Kulesza, A.; Pereira, F.; and
  Vaughan, J.~W.
\newblock 2010.
\newblock A theory of learning from different domains.
\newblock {\em Machine Learning} 79(1-2):151--175.

\bibitem[\protect\citeauthoryear{Berthelot \bgroup et al\mbox.\egroup
  }{2019}]{autoencoder_interpolation}
Berthelot, D.; Raffel, C.; Roy, A.; and Goodfellow, I.
\newblock 2019.
\newblock Understanding and improving interpolation in autoencoders via an
  adversarial regularizer.
\newblock In {\em ICLR}.

\bibitem[\protect\citeauthoryear{Bousmalis \bgroup et al\mbox.\egroup
  }{2017}]{pixel_level}
Bousmalis, K.; Silberman, N.; Dohan, D.; Erhan, D.; and Krishnan, D.
\newblock 2017.
\newblock Unsupervised pixel-level domain adaptation with generative
  adversarial networks.
\newblock In {\em CVPR}.

\bibitem[\protect\citeauthoryear{Ganin and Lempitsky}{2015}]{gradient_reverse}
Ganin, Y., and Lempitsky, V.~S.
\newblock 2015.
\newblock Unsupervised domain adaptation by backpropagation.
\newblock In {\em ICML}.

\bibitem[\protect\citeauthoryear{Ganin \bgroup et al\mbox.\egroup
  }{2016}]{domain_adversarial_journal}
Ganin, Y.; Ustinova, E.; Ajakan, H.; Germain, P.; Larochelle, H.; Laviolette,
  F.; Marchand, M.; and Lempitsky, V.
\newblock 2016.
\newblock Domain-adversarial training of neural networks.
\newblock {\em JMLR} 17(1):2096--2030.

\bibitem[\protect\citeauthoryear{Ghifary \bgroup et al\mbox.\egroup
  }{2016}]{deep_reconstruction}
Ghifary, M.; Kleijn, W.~B.; Zhang, M.; Balduzzi, D.; and Li, W.
\newblock 2016.
\newblock Deep reconstruction-classification networks for unsupervised domain
  adaptation.
\newblock In {\em ECCV}.

\bibitem[\protect\citeauthoryear{Goodfellow \bgroup et al\mbox.\egroup
  }{2014}]{gan}
Goodfellow, I.~J.; Pouget{-}Abadie, J.; Mirza, M.; Xu, B.; Warde{-}Farley, D.;
  Ozair, S.; Courville, A.~C.; and Bengio, Y.
\newblock 2014.
\newblock Generative adversarial nets.
\newblock In {\em NeurIPS}.

\bibitem[\protect\citeauthoryear{Gretton \bgroup et al\mbox.\egroup
  }{2012}]{kernel_two_sample}
Gretton, A.; Borgwardt, K.~M.; Rasch, M.~J.; Sch{\"{o}}lkopf, B.; and Smola,
  A.~J.
\newblock 2012.
\newblock A kernel two-sample test.
\newblock {\em JMLR} 13:723--773.

\bibitem[\protect\citeauthoryear{He \bgroup et al\mbox.\egroup }{2016}]{resnet}
He, K.; Zhang, X.; Ren, S.; and Sun, J.
\newblock 2016.
\newblock Deep residual learning for image recognition.
\newblock In {\em CVPR}.

\bibitem[\protect\citeauthoryear{Hoffman \bgroup et al\mbox.\egroup
  }{2018}]{cycada}
Hoffman, J.; Tzeng, E.; Park, T.; Zhu, J.; Isola, P.; Saenko, K.; Efros, A.~A.;
  and Darrell, T.
\newblock 2018.
\newblock Cycada: Cycle-consistent adversarial domain adaptation.
\newblock In {\em ICML}.

\bibitem[\protect\citeauthoryear{Hull}{1994}]{usps}
Hull, J.~J.
\newblock 1994.
\newblock A database for handwritten text recognition research.
\newblock {\em {IEEE} Trans. Pattern Anal. Mach. Intell.} 16(5):550--554.

\bibitem[\protect\citeauthoryear{Kang \bgroup et al\mbox.\egroup
  }{2018}]{adversarial_attention}
Kang, G.; Zheng, L.; Yan, Y.; and Yang, Y.
\newblock 2018.
\newblock Deep adversarial attention alignment for unsupervised domain
  adaptation: The benefit of target expectation maximization.
\newblock In {\em ECCV}.

\bibitem[\protect\citeauthoryear{Kingma and Welling}{2013}]{vae}
Kingma, D.~P., and Welling, M.
\newblock 2013.
\newblock Auto-encoding variational bayes.
\newblock {\em CoRR} abs/1312.6114.

\bibitem[\protect\citeauthoryear{Krizhevsky, Sutskever, and
  Hinton}{2012}]{alexnet}
Krizhevsky, A.; Sutskever, I.; and Hinton, G.~E.
\newblock 2012.
\newblock Imagenet classification with deep convolutional neural networks.
\newblock In {\em NeurIPS}.

\bibitem[\protect\citeauthoryear{Larsen \bgroup et al\mbox.\egroup
  }{2016}]{vae-gan}
Larsen, A. B.~L.; S{\o}nderby, S.~K.; Larochelle, H.; and Winther, O.
\newblock 2016.
\newblock Autoencoding beyond pixels using a learned similarity metric.
\newblock In {\em ICML}.

\bibitem[\protect\citeauthoryear{Lin \bgroup et al\mbox.\egroup }{2014}]{coco}
Lin, T.; Maire, M.; Belongie, S.~J.; Hays, J.; Perona, P.; Ramanan, D.;
  Doll{\'{a}}r, P.; and Zitnick, C.~L.
\newblock 2014.
\newblock Microsoft {COCO:} common objects in context.
\newblock In {\em ECCV}.

\bibitem[\protect\citeauthoryear{Liu and Tuzel}{2016}]{co-gan}
Liu, M., and Tuzel, O.
\newblock 2016.
\newblock Coupled generative adversarial networks.
\newblock In {\em NeurIPS}.

\bibitem[\protect\citeauthoryear{Long \bgroup et al\mbox.\egroup
  }{2015}]{Long2015}
Long, M.; Cao, Y.; Wang, J.; and Jordan, M.~I.
\newblock 2015.
\newblock Learning transferable features with deep adaptation networks.
\newblock In {\em ICML}.

\bibitem[\protect\citeauthoryear{Long \bgroup et al\mbox.\egroup
  }{2017}]{Long2017}
Long, M.; Zhu, H.; Wang, J.; and Jordan, M.~I.
\newblock 2017.
\newblock Deep transfer learning with joint adaptation networks.
\newblock In {\em ICML}.

\bibitem[\protect\citeauthoryear{Long \bgroup et al\mbox.\egroup }{2018}]{cdan}
Long, M.; Cao, Z.; Wang, J.; and Jordan, M.~I.
\newblock 2018.
\newblock Conditional adversarial domain adaptation.
\newblock In {\em NeurIPS}.

\bibitem[\protect\citeauthoryear{Lécun \bgroup et al\mbox.\egroup
  }{1998}]{mnist}
Lécun, Y.; Bottou, L.; Bengio, Y.; and Haffner, P.
\newblock 1998.
\newblock Gradient-based learning applied to document recognition.
\newblock {\em Proceedings of the IEEE} 86(11):2278--2324.

\bibitem[\protect\citeauthoryear{Ma, Zhang, and Xu}{2019}]{gcan}
Ma, X.; Zhang, T.; and Xu, C.
\newblock 2019.
\newblock Gcan: Graph convolutional adversarial network for unsupervised domain
  adaptation.
\newblock In {\em CVPR}.

\bibitem[\protect\citeauthoryear{Maaten and Hinton}{2008}]{tsne}
Maaten, L. V.~D., and Hinton, G.
\newblock 2008.
\newblock Visualizing data using t-sne.
\newblock {\em JMLR} 9(2605):2579--2605.

\bibitem[\protect\citeauthoryear{Mansour, Mohri, and
  Rostamizadeh}{2009}]{A_distance_2}
Mansour, Y.; Mohri, M.; and Rostamizadeh, A.
\newblock 2009.
\newblock Domain adaptation: Learning bounds and algorithms.
\newblock In {\em Annual Conference on Learning Theory}.

\bibitem[\protect\citeauthoryear{Netzer \bgroup et al\mbox.\egroup
  }{2011}]{svhn}
Netzer, Y.; Wang, T.; Coates, A.; Bissacco, A.; Wu, B.; and Ng, A.~Y.
\newblock 2011.
\newblock Reading digits in natural images with unsupervised feature learning.
\newblock {\em NeurIPS Workshop}.

\bibitem[\protect\citeauthoryear{Pan and Yang}{2010}]{survey}
Pan, S.~J., and Yang, Q.
\newblock 2010.
\newblock A survey on transfer learning.
\newblock {\em {IEEE} Trans. Knowl. Data Eng.} 22(10):1345--1359.

\bibitem[\protect\citeauthoryear{Pan \bgroup et al\mbox.\egroup }{2011}]{tca}
Pan, S.~J.; Tsang, I.~W.; Kwok, J.~T.; and Yang, Q.
\newblock 2011.
\newblock Domain adaptation via transfer component analysis.
\newblock {\em {IEEE} Trans. Neural Networks} 22(2):199--210.

\bibitem[\protect\citeauthoryear{Paszke \bgroup et al\mbox.\egroup
  }{2017}]{pytorch}
Paszke, A.; Gross, S.; Chintala, S.; Chanan, G.; Yang, E.; DeVito, Z.; Lin, Z.;
  Desmaison, A.; Antiga, L.; and Lerer, A.
\newblock 2017.
\newblock Automatic differentiation in pytorch.
\newblock In {\em NeurIPS Workshop}.

\bibitem[\protect\citeauthoryear{Pei \bgroup et al\mbox.\egroup
  }{2018}]{multi-adversarial}
Pei, Z.; Cao, Z.; Long, M.; and Wang, J.
\newblock 2018.
\newblock Multi-adversarial domain adaptation.
\newblock In {\em AAAI}.

\bibitem[\protect\citeauthoryear{Peng \bgroup et al\mbox.\egroup
  }{2017}]{visda}
Peng, X.; Usman, B.; Kaushik, N.; Hoffman, J.; Wang, D.; and Saenko, K.
\newblock 2017.
\newblock Visda: The visual domain adaptation challenge.
\newblock {\em CoRR} abs/1710.06924.

\bibitem[\protect\citeauthoryear{Russakovsky \bgroup et al\mbox.\egroup
  }{2015}]{imagenet}
Russakovsky, O.; Deng, J.; Su, H.; Krause, J.; Satheesh, S.; Ma, S.; Huang, Z.;
  Karpathy, A.; Khosla, A.; Bernstein, M.~S.; Berg, A.~C.; and Li, F.
\newblock 2015.
\newblock Imagenet large scale visual recognition challenge.
\newblock {\em IJCV} 115(3):211--252.

\bibitem[\protect\citeauthoryear{Saenko \bgroup et al\mbox.\egroup
  }{2010}]{office}
Saenko, K.; Kulis, B.; Fritz, M.; and Darrell, T.
\newblock 2010.
\newblock Adapting visual category models to new domains.
\newblock In {\em ECCV}.

\bibitem[\protect\citeauthoryear{Saito \bgroup et al\mbox.\egroup
  }{2018a}]{adversarial_dropout}
Saito, K.; Ushiku, Y.; Harada, T.; and Saenko, K.
\newblock 2018a.
\newblock Adversarial dropout regularization.
\newblock In {\em ICLR}.

\bibitem[\protect\citeauthoryear{Saito \bgroup et al\mbox.\egroup
  }{2018b}]{max_discrepancy}
Saito, K.; Watanabe, K.; Ushiku, Y.; and Harada, T.
\newblock 2018b.
\newblock Maximum classifier discrepancy for unsupervised domain adaptation.
\newblock In {\em CVPR}.

\bibitem[\protect\citeauthoryear{Sankaranarayanan \bgroup et al\mbox.\egroup
  }{2018}]{generate_to_adapt}
Sankaranarayanan, S.; Balaji, Y.; Castillo, C.~D.; and Chellappa, R.
\newblock 2018.
\newblock Generate to adapt: Aligning domains using generative adversarial
  networks.
\newblock In {\em CVPR}.

\bibitem[\protect\citeauthoryear{Sun and Saenko}{2016}]{deep_coral}
Sun, B., and Saenko, K.
\newblock 2016.
\newblock Deep coral: Correlation alignment for deep domain adaptation.
\newblock In {\em ECCV Workshop}.

\bibitem[\protect\citeauthoryear{Tran \bgroup et al\mbox.\egroup
  }{2019}]{joint_pixel_feature}
Tran, L.; Sohn, K.; Yu, X.; Liu, X.; and Chandraker, M.
\newblock 2019.
\newblock Gotta adapt 'em all: Joint pixel and feature-level domain adaptation
  for recognition in the wild.
\newblock In {\em CVPR}.

\bibitem[\protect\citeauthoryear{Tzeng \bgroup et al\mbox.\egroup }{2014}]{ddc}
Tzeng, E.; Hoffman, J.; Zhang, N.; Saenko, K.; and Darrell, T.
\newblock 2014.
\newblock Deep domain confusion: Maximizing for domain invariance.
\newblock {\em CoRR} abs/1412.3474.

\bibitem[\protect\citeauthoryear{Tzeng \bgroup et al\mbox.\egroup
  }{2017}]{adda}
Tzeng, E.; Hoffman, J.; Saenko, K.; and Darrell, T.
\newblock 2017.
\newblock Adversarial discriminative domain adaptation.
\newblock In {\em CVPR}.

\bibitem[\protect\citeauthoryear{Xie \bgroup et al\mbox.\egroup
  }{2018}]{semantic}
Xie, S.; Zheng, Z.; Chen, L.; and Chen, C.
\newblock 2018.
\newblock Learning semantic representations for unsupervised domain adaptation.
\newblock In {\em ICML}.

\bibitem[\protect\citeauthoryear{Yan \bgroup et al\mbox.\egroup }{2017}]{wdan}
Yan, H.; Ding, Y.; Li, P.; Wang, Q.; Xu, Y.; and Zuo, W.
\newblock 2017.
\newblock Mind the class weight bias: Weighted maximum mean discrepancy for
  unsupervised domain adaptation.
\newblock In {\em CVPR}.

\bibitem[\protect\citeauthoryear{Yosinski \bgroup et al\mbox.\egroup
  }{2014}]{deep_transfer_ability}
Yosinski, J.; Clune, J.; Bengio, Y.; and Lipson, H.
\newblock 2014.
\newblock How transferable are features in deep neural networks?
\newblock In {\em NeurIPS}.

\bibitem[\protect\citeauthoryear{Zhang \bgroup et al\mbox.\egroup
  }{2018a}]{mixup}
Zhang, H.; Ciss{\'{e}}, M.; Dauphin, Y.~N.; and Lopez{-}Paz, D.
\newblock 2018a.
\newblock mixup: Beyond empirical risk minimization.
\newblock In {\em ICLR}.

\bibitem[\protect\citeauthoryear{Zhang \bgroup et al\mbox.\egroup
  }{2018b}]{collaborative}
Zhang, W.; Ouyang, W.; Li, W.; and Xu, D.
\newblock 2018b.
\newblock Collaborative and adversarial network for unsupervised domain
  adaptation.
\newblock In {\em CVPR}.

\end{thebibliography}

\end{document}